\documentclass[runningheads]{llncs}
\usepackage{graphicx}

\usepackage{tikz}
\usepackage{comment}
\usepackage{amsmath,amssymb} 
\usepackage{color}

\usepackage[accsupp]{axessibility}  

\usepackage{times}
\usepackage{epsfig}
\usepackage{caption}
\usepackage{subcaption}
\usepackage{tabularx}
\usepackage{makecell}
\usepackage{cellspace}
\usepackage{graphicx}
\usepackage{wrapfig}

\makeatletter
\@namedef{ver@everyshi.sty}{}
\makeatother
\usepackage{tikz}
\usepackage{pgfplots}
\pgfplotsset{compat=1.17}
\usetikzlibrary{pgfplots.groupplots}

\usepackage{algorithm}
\usepackage{algpseudocode}

\newcommand{\methodname}{\emph{GraphVid}}

\DeclareMathOperator*{\argmin}{argmin}

\def\Real{\mathbb{R}}
\def\neighborhood{\mathcal{N}}
\def\mathg{\mathcal{G}}
\def\mathv{\mathcal{V}}
\def\mathe{\mathcal{E}}
\def\mathr{\mathcal{R}}

\def\eg{\emph{e.g.}} 
\def\ie{\emph{i.e.}}

\begin{document}
\pagestyle{headings}
\mainmatter
\def\ECCVSubNumber{4861}  

\title{\methodname:\ It Only Takes a Few Nodes to Understand a Video} 


\titlerunning{\methodname:\ It Only Takes a Few Nodes to Understand a Video}
%
\author{Eitan Kosman\orcidID{0000-0002-5538-0616} \and
Dotan Di Castro}
\authorrunning{E. Kosman and D. Di Castro}
%
\institute{Bosch Center of AI, Haifa, Israel\\
\email{\{Eitan.Kosman,Dotan.DiCastro\}@bosch.com}}
\maketitle

\begin{abstract}\label{section:abstract}
We propose a concise representation of videos that encode perceptually meaningful features into graphs. With this representation, we aim to leverage the large amount of redundancies in videos and save computations. First, we construct superpixel-based graph representations of videos by considering superpixels as graph nodes and create spatial and temporal connections between adjacent superpixels. Then, we leverage Graph Convolutional Networks to process this representation and predict the desired output. As a result, we are able to train models with much fewer parameters, which translates into short training periods and a reduction in computation resource requirements. A comprehensive experimental study on the publicly available datasets Kinetics-400 and Charades shows that the proposed method is highly cost-effective and uses limited commodity hardware during training and inference. \textbf{It reduces the computational requirements 10-fold} while achieving results that are comparable to state-of-the-art methods. We believe that the proposed approach is a promising direction that could open the door to solving video understanding more efficiently and enable more resource limited users to thrive in this research field.
\end{abstract}

\section{Introduction}\label{section:introduction}
The field of video understanding has gained prominence thanks to the rising popularity of videos, which has become the most common form of data on the web. On each new uploaded video, a variety of tasks can be performed, such as tagging \cite{fernandez2017vits}, human action recognition \cite{pareek2021survey}, anomaly detection \cite{suarez2020survey}, etc. New video-processing algorithms are continuously being developed to automatically organize the web through the flawless accomplishment of the aforementioned tasks.


Nowadays, Deep Neural Networks are the de-facto standard for video understanding \cite{oprea2020review}. However, with every addition of a new element to the training set (that is, a full training video), more resources are required in order to satisfy the enormous computational needs.
On the one hand, the exponential increment in the amount of data raises concerns regarding our ability to handle it in the future. On the other hand, it has also spurred an highly creative research field aimed at finding ways to mitigate this burden. 

Among the first-generation of video processing methods were ones geared toward adopting 2D convolution neural networks (CNNs), due to their computational efficiency \cite{simonyan2014two}. Others decomposed 3D convolutions \cite{du2017closer,xie2018rethinking} into simpler operators, or split a complex neural network into an ensemble of
lightweight networks \cite{chen2018multi}. However, video understanding has greatly evolved since then, with the current state-of-the-art methods featuring costly attention mechanisms \cite{arnab2021vivit,girdhar2019video,liu2021video,akbari2021vatt,fan2021multiscale,bertasius2021space,li2021vidtr}. Beyond accuracy, a prominent advantage of the latest generation of methods is that they process raw data, that is, video frames that do not undergo any advanced pre-processing. Meanwhile, pursuing new video representations and incorporating pre-computed features to accelerate training is a promising direction that requires more extensive research.
\newcommand{\thumbwidth}{0.2}
\newcommand{\thumbheight}{1.2in}
\begin{figure}[ht]
    \centering
    \begin{subfigure}[b]{0.4\linewidth}
         \centering
         \includegraphics[width=0.4\linewidth]{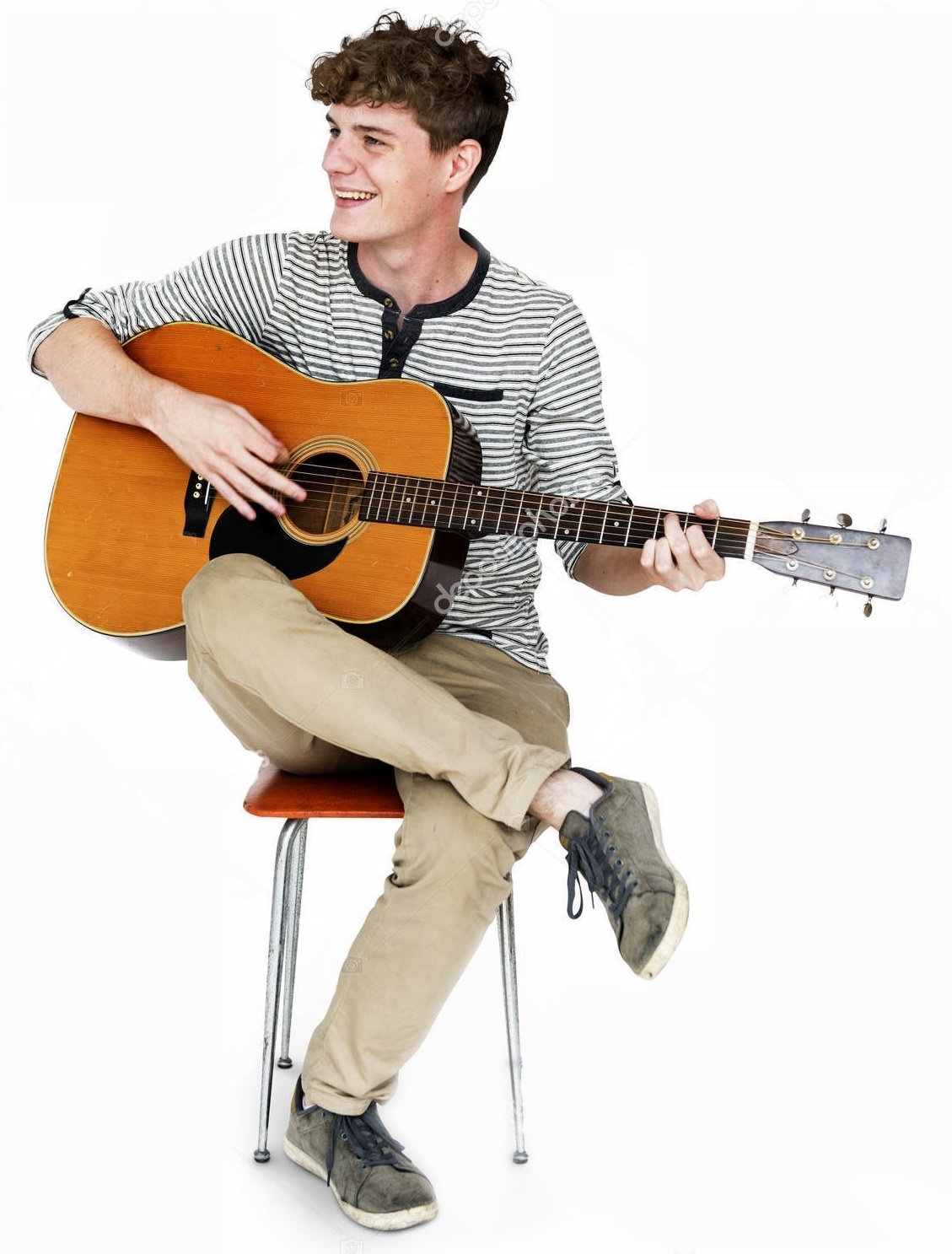}
        \caption{Original image}
         \label{fig:original_intro}
     \end{subfigure}
    \begin{subfigure}[b]{0.4\linewidth}
         \centering
         \includegraphics[width=0.4\linewidth]{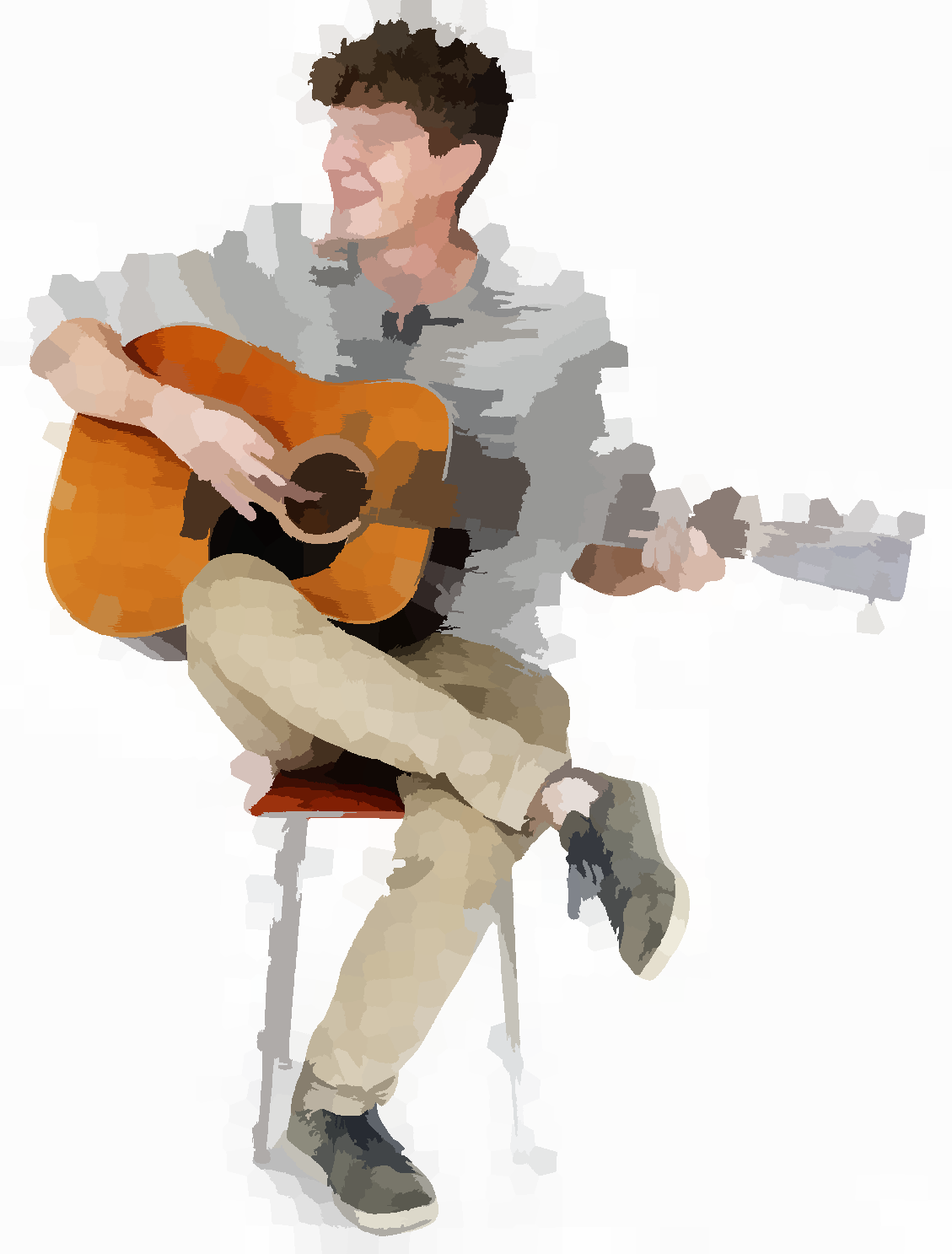}
        \caption{Mean superpixels}
         \label{fig:superpixels_intro}
     \end{subfigure}
    \caption{A visual comparison between a pixel and a mean-superpixel representation. On the left, the original image is presented. On the right, we present the image formed by generating superpixel regions using SLIC and filling each region with its mean color.}
    \label{fig:superpixels_example}
\end{figure}

Prior to the renaissance of deep learning \cite{lecun2015deep}, much research was done on visual feature generation. Two prominent visual feature generation methods are superpixels\footnote{Superpixel techniques segment an image into regions by considering similarity measures, defined using perceptual features.} and optic-flow\footnote{Optic-flow is the pattern of the apparent motion of an object(s) in the image between two consecutive frames due to the movement of the object or the camera.}. These techniques' ability to encode perceptually meaningful features has greatly contributed to the success of computer vision algorithms. Superpixels provide a convenient, compact representation of images that can be very useful for computationally demanding problems, while optic-flow provides hints about motion. We rely on these methods to construct a novel representation of videos that encodes sufficient information for video understanding: 1) adjacent pixels are grouped together in the form of superpixels, and 2) temporal relations and proximities are expressed via graph connectivity. The example depicted in Figure \ref{fig:superpixels_example} provides an intuition for the sufficiency of superpixel representation for scene understanding. It contains the superpixel regions obtained via SLIC \cite{achanta2010slic}, with each region filled with the mean color. One can clearly discern a person playing a guitar in both images. A different way of depicting the relations between superpixels is a graph with nodes representing superpixels \cite{monti2017geometric,dadsetan2021superpixels,avelar2020superpixel}. Such a representation has the advantage of being invariant to rotations and flips, which obviates the need for further augmentations. We here demonstrate how this representation can reduce the computational requirements for processing videos.

Recent years have seen a surge in the utilization of Graph Neural Networks (GNNs) \cite{kipf2016semi} in tasks that involve images \cite{monti2017geometric,dadsetan2021superpixels,avelar2020superpixel}, audio \cite{dokania2019graph,zhang2019few} and other data forms \cite{wang2018videos,xie2016representation,abadal2021computing}. In this paper, we propose \methodname, a concise graph representation of videos that enables video processing via GNNs. \methodname\ constructs a graph representation of videos that is subsequently processed via a GCN to predict a target. We intend to exploit the power of graphs for efficient video processing. To the best of our knowledge, we are the first to utilize a graph-based representation of videos for efficiency. \methodname\ dramatically reduces the memory footprint of a model, enabling large batch-sizes that translate to better generalization. Moreover, it utilizes models with an order-of-magnitude fewer parameters than the current state-of-the-art models while preserving the predictive power. \textbf{In summary, our contributions are:}
\begin{enumerate}
    \item We present \methodname\ - a simple and intuitive, yet sufficient representation of video clips. This simplicity is crucial for delivering efficiency.
    \item We propose a dedicated GNN for processing the proposed representation. The proposed architecture is compared with conventional GNN models in order to demonstrate the importance of each component of \methodname.
    \item We present 4 types of new augmentations that are directly applied to the video-graph representation. A thorough ablation study of their configurations is preformed in order to demonstrate the contribution of each.
    \item We perform a thorough experimental study, and show that \methodname\ greatly outperforms previous methods in terms of efficiency - first and foremost, the paper utilizes GNNs for efficient video understanding. We show that it successfully
reduces computations while preserving much of the performance of state-of-the-art approaches that utilize computationally demanding models.
\end{enumerate}
\section{Related Work}\label{section:related_work}

\subsection{Deep Learning for Video Understanding}
CNNs have found numerous applications in video processing \cite{mittal2021survey,tran2018closer,yue2015beyond}. These include LSTM-based networks that perform per-frame encoding \cite{srivastava2015unsupervised,ullah2017action,yue2015beyond} and the extension of 2D convolutions to the temporal dimension, \eg, 3D
CNNs such as C3D \cite{tran2015learning}, R2D \cite{simonyan2014two} and R(2+1)D \cite{tran2018closer}.

The success of the Transformer model \cite{vaswani2017attention} has led to the development of attention-based models for vision tasks, via self-attention modules that were used to model spatial dependencies in images. NLNet \cite{wang2018non} was the first to employ self-attention in a CNN. With this novel attention mechanism, NLNet is possible to model long-range dependencies between pixels. The next model to be developed was GCNet \cite{cao2019gcnet}, which simplified the NL-module, thanks to its need for fewer parameters and computations, while preserving its performance. A more prominent transition from CNNs to Transformers began with Vision Transformer (ViT) \cite{dosovitskiy2020image}, which prompted research aimed at improving its effectiveness on small datasets, such as Deit \cite{touvron2021training}. Later, vision-transformers were adapted for video tasks \cite{neimark2021video,arnab2021vivit,bertasius2021space,fan2021multiscale,li2021vidtr,liu2021video}, now crowned as the current state-of-the-art that top the leader-boards of this field. 

The usage of graph representation in video understanding sparsely took place in the work of Wang \cite{wang2018videos}. They used pre-trained Resnet variants \cite{he2016deep} for generating object bounding boxes of interest on each frame. These bounding boxes are later used for the construction of a spatio-temporal graph that describes how objects change through time, and perform classification on top of the spatio-temporal graph with graph convolutional neural networks \cite{kipf2016semi}. However, we note that the usage of a large backbone for generating object bounding boxes is harmful for performance. We intend to alleviate this by proposing a lighter graph representation. In combination of a dedicated GNN architecture, our representation greatly outperforms \cite{wang2018videos} in all metrics.

\subsection{Superpixel Representation of Visual Data}
Superpixels are groups of perceptually similar pixels that can be used to create visually meaningful entities while heavily reducing the number of primitives for subsequent processing steps \cite{stutz2018superpixels}. The efficiency of the obtained representation has led to the development of many superpixel-generation algorithms for images \cite{stutz2018superpixels}.
This approach was adapted for volumetric data via the construction of supervoxels \cite{papon2013voxel}, which are the trivial extension to depth. These methods were adjusted for use in videos \cite{6247802} by treating the temporal dimension as depth. However, this results in degraded performance, as inherent assumptions regarding neighboring points in the 3D space do not apply to videos with non-negligible motion. Recent approaches especially designed to deal with videos consider the temporal dimensions for generating superpixels that are coherent in time. Xu \emph{et al.}~\cite{10.1007/978-3-642-33783-3_45} proposed a hierarchical graph-based segmentation method. This was followed by the work of Chang \emph{et al.}~\cite{chang2013video}, who suggested that Temporal Superpixels (TSPs) can serve as a representation of videos using temporal superpixels by modeling the flow between frames with a bilateral Gaussian process.

\subsection{Graph Convolutional Neural Networks}
Introduced in \cite{kipf2016semi}, Graph Convolutional Networks (GCNs) have been widely adopted for graph-related tasks \cite{zhang2018network,kumar2020link}.
The basic GCN uses aggregators, such as average and summation, to obtain a node representation given its neighbors. This basic form was rapidly extended to more complex architectures with more sophisticated aggregators. For instance, Graph Attention Networks \cite{velivckovic2017graph} use dot-product-based attention to calculate weights for edges. 
Relational GCNs \cite{schlichtkrull2018modeling} add to this framework by also considering multiple edge types, namely, relations (such as temporal and spatial relations), and the aggregating information from each relation via separate weights in a single layer.
Recently, GCNs have been adopted for tasks involving audio \cite{dokania2019graph,zhang2019few} and images \cite{monti2017geometric,dadsetan2021superpixels,avelar2020superpixel}. Following the success of graph models to efficiently perform image-based tasks, we are eager to demonstrate our extension of the image-graph representation to videos.
\section{\methodname\ - A Video-Graph Representation}\label{section:methodology}
\begin{figure*}[!ht]
    \centering
    \includegraphics[width=0.8\linewidth]{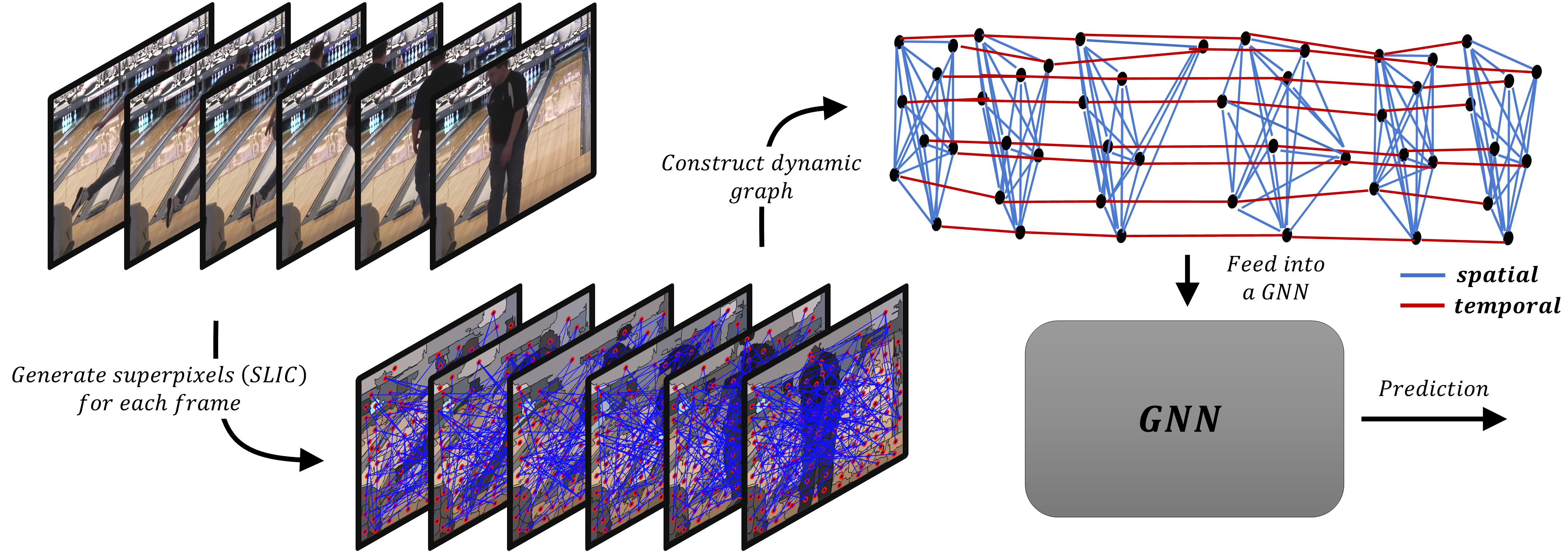}
    \caption{The flow of \methodname. Given a video clip, we generate superpixels using SLIC for each frame. The superpixels are used to construct a region-adjacency graph of a frame, with superpixels as nodes. Then, the graph sequence is connected via temporal proximities to construct a dynamic graph, which is later fed into a GNN for prediction.} 
    \label{fig:framework}
\end{figure*}

In this section, we introduce the methodology of \methodname. First, we present our method for video-graph representation generation, depicted in Figure \ref{fig:framework} and described in Algorithm \ref{algo:graphvid}. Then, we present our training methodology that utilizes this representation. Finally, we discuss the benefits of \methodname\ and propose several augmentations.

\subsection{Overview}
In our framework, we deal with video clips that are sequences of $T$ video frames \text{$v\in \Real^{T\times~C\times~H\times~W}$}. The goal is to transform $v$ into a graph that is sufficiently informative for further processing. To achieve this, we use SLIC \cite{achanta2010slic} to generate $S$ segmented regions, called \textit{superpixels}, over each frame. We denote each segmented region as $R_{t,i}$, where \text{$t\in [T]$} represents the temporal frame index, and \text{$i\in [S]$} the superpixel-segmented region index. The following is a description of how we utilize the superpixels to construct our video-graph representation.

\paragraph{Graph Elements -}
We define the undirected graph $\mathg$ as a 3-tuple \text{$\mathg=(\mathv,\mathe,\mathr)$}, where \text{$\mathv=\{R_{t,i} | t\in [T], i\in [S]\}$} is the set of nodes representing the segmented regions, $\mathe$ is the set of labeled edges (to be defined hereunder) and \text{$\mathr=\{spatial,temporal\}$} is a set of relations as defined in \cite{schlichtkrull2018modeling}. Each node $R_{t,i}$ is associated with an attribute $R_{t,i}.c\in \Real^3$ representing the mean RGB color in that segmented region. Additionally, we refer to $R_{t,i}.y$ and $R_{t,i}.x$ as the coordinates of the superpixel's centroid, which we use to compute the distances between superpixels. These distances, which will later serve as the edge attributes of the graph, are computed by
\begin{equation}
    d^{t_q\to t_p}_{i,j} = \sqrt{\left(\frac{R_{t_q,i}.y - R_{t_p,j}.y}{H}\right)^2 + \left(\frac{R_{t_q,i}.x - R_{t_p,j}.x}{W}\right)^2}.
\end{equation}
Here, \text{$t_q,t_p\in [T]$} denote frame indices, and \text{$i,j\in [S]$} denote superpixel indices generated for the corresponding frames.
The set of edges $\mathe$ is composed of: \textbf{1)} intra-frame edges (denoted $\mathe^{spatial}$) - edges between nodes corresponding to superpixels in the same frame. We refer to these as \textit{spatial edges}. \textbf{2)} inter-frame edges (denoted $\mathe^{temporal}$) - edges between nodes corresponding to superpixels in two sequential frames. We refer to edges as \textit{temporal edges}.
Finally, the full set of edges is \text{$\mathe = \mathe^{spatial} \cup \mathe^{temporal}$}.
Following is a description of how we construct both components.

\begin{figure}[!ht]
    \centering
    \includegraphics[width=0.45\linewidth]{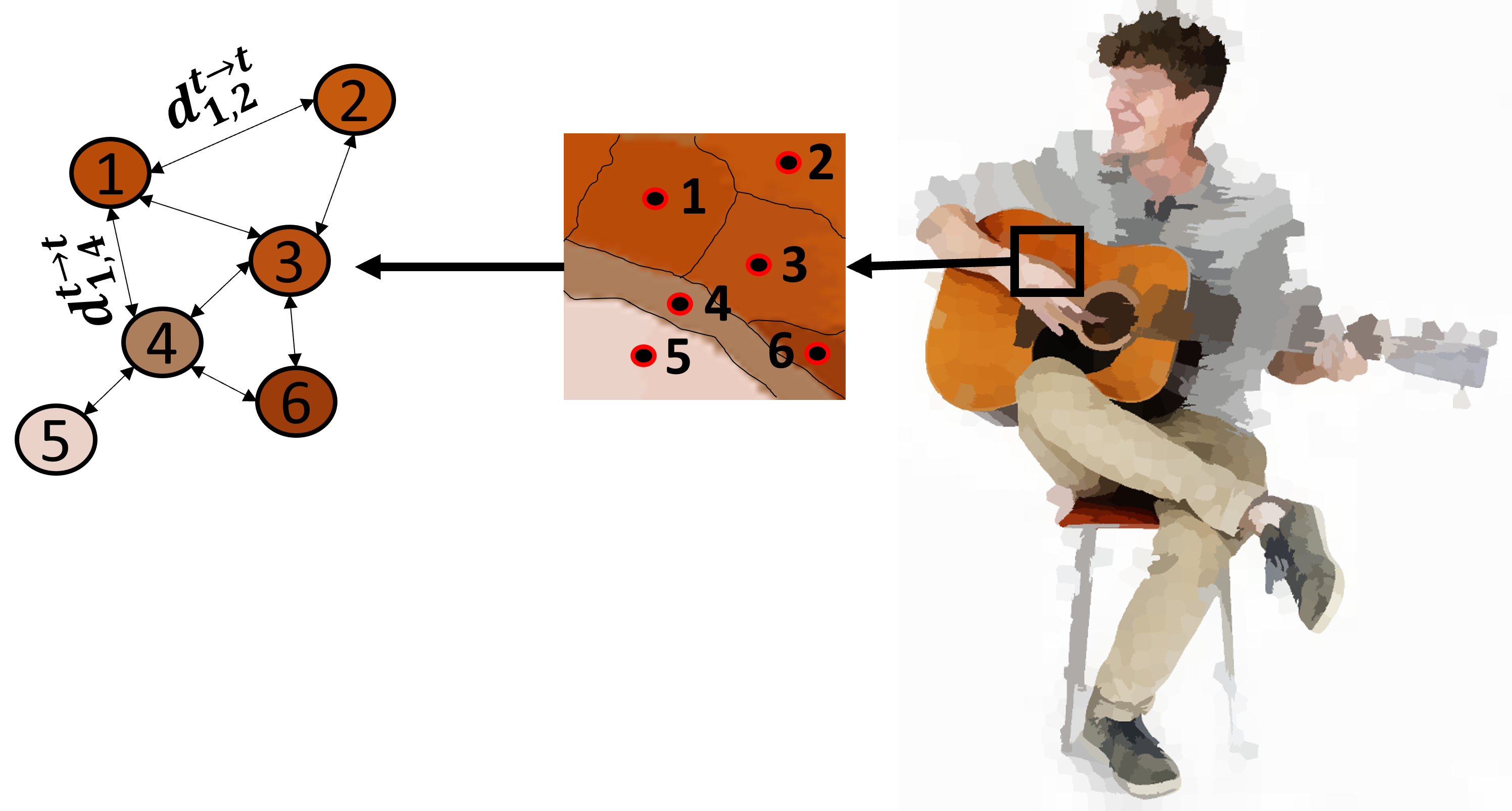}
    \caption{Spatial edge generation. First, superpixels are generated. Each superpixel is represented as a node, which is connected via its edges to other such nodes within a frame. Each node is assigned the mean color of the respective segmented region, and each edge is assigned the distances between the superpixel centroids connected by that edge.}
    \label{fig:spatial_edges}
\end{figure}

\paragraph{Spatial Edges -}
In similar to \cite{avelar2020superpixel}, we generate a region-adjacency graph for each frame, with edge attributes describing the distances between superpixel centroids. The notation \text{$\mathe^{spatial}_t$} refers to the set of the spatial-edges connecting nodes corresponding to superpixels in the frame $t$, 
and
\(
    \mathe^{spatial} = \bigcup_{t=1}^{T}{\mathe^{spatial}_t}.
\)
Each edge \text{$e_{i,j}^{t}\in \mathe^{spatial}$} is associated with an attribute that describes the euclidean distance between the two superpixel centroids $i$ and $j$ in frame $t$, that is, $d^{t\to t}_{i,j}$.
These distances provide information about the relations between the superpixels. Additionally, the distances are invariant to rotations and image-flips, which eliminates the need for those augmentations. Note that normalization of the superpixels' centroid coordinates is required in order to obscure information regarding the resolution of frames, which is irrelevant for many tasks, such as action classification. In Figure \ref{fig:spatial_edges}, we demonstrate the procedure of spatial edge generation for a cropped image that results in a partial graph of the whole image. Each superpixel is associated with a node, which is connected via edges to other adjacent nodes (with the distances between the superpixels' centroids serving as edge attributes). 

\paragraph{Temporal Edges -}
In modeling the temporal relations, we aim to connect nodes that tend to describe the same objects in subsequent frames. To do so, we rely on the assumption that in subsequent frames, such superpixels are attributed similar colors and the same spatial proximity. To achieve this, for each superpixel $R_{t,i}$, we construct a neighborhood $\neighborhood_{t,i}$ that contains superpixels from its subsequent frame whose centroids have a proximity of at most $d_{proximity}\in (0,1]$ with respect to the euclidean distance. Then, we find the superpixel with the most similar color in this neighborhood. As a result, the $t^{th}$ frame is associated with the set of edges $\mathe^{temporal}_{t\to t+1}$ that model temporal relations with its subsequent frame, formally:
\begin{equation}\label{eq:neighborhood}
    \neighborhood_{t,i} = \{R_{t+1,j} | d^{t\to t+1}_{i,j} < d_{proximity}\},
\end{equation}
\begin{equation}
    neighbor(R_{t,i})=\argmin_{R_{t+1,j}\in \neighborhood_{t,i}}{|R_{t,i}.c - R_{t+1,j}.c|_2},
\end{equation}
\begin{equation}
    \mathe^{temporal}_{t\to t+1} = \{(R_{t,i}, temporal, neighbor(R_{t,i}) | i\in [S]\}.
\end{equation}

Equipped with these definitions, we define the set of temporal edges connecting nodes corresponding to superpixels in frame $t$ to superpixels in frame \text{$t+1$} as the union of the temporal edge sets generated for all the frames:
\(
    \mathe^{temporal} = \bigcup_{t=1}^{T-1}{\mathe^{temporal}_{t\to t+1}}
\).
\begin{algorithm}[!ht]
    \caption{Graph Generation}
    \label{algo:graphvid}
    \begin{algorithmic}
    \State \textbf{Input:} $v\in \Real^{T\times C\times H \times W}$ \Comment{The input video clip}
    \State \textbf{Parameters:}
    $S\in \mathbb{N}$ \Comment{Number of superpixels per frame}
    \State \ \ \ \ \ \ \ \ \ \ \ \ \ \ \ \ \ \ \ \ \ \ $d_{proximity}\in (0,1]$ \Comment{Diameter of neighborhoods}
    \State \textbf{Output:} $\mathg=(\mathv,\mathe,\mathr)$ \Comment{A video-graph}
    \State $\mathv,\ \mathv_{last},\ \mathe^{spatial},\ \mathe^{temporal} \gets \emptyset, \emptyset, \emptyset, \emptyset$
    \For{$t\in [T]$}
        \State $SP \gets SLIC(v[t], S)$
        \State $\mathv \gets \mathv \cup SP$
        \State $\mathe^{spatial} \gets \mathe^{spatial} \cup regionAdjacetEdges(SP)$
        \State $\mathe^{temporal}_{t-1\to t} \gets \emptyset$
        \For{$R_{t-1, i}\in \mathv_{last}$}
            \State $\neighborhood_{t-1,i} \gets \{R_{t,j} | d^{t-1\to t}_{i,j} < d_{proximity}\}$
            \State $nn_{t-1,i} \gets \argmin_{R_{t,j}\in \neighborhood_{t,i}}{|R_{t,i}.c - R_{t,j}.c|_2})$
            \State $\mathe^{temporal}_{t-1\to t} \gets \mathe^{temporal}_{t-1\to t} \cup \{(R_{t-1, i}, temporal, nn_{t-1,i})\}$
        \EndFor
        \State $\mathe^{temporal} \gets \mathe^{temporal} \cup \mathe^{temporal}_{t-1\to t}$
        \State $\mathv_{last} \gets SP$
    \EndFor
    \State \Return $\mathg=(\mathv,\mathe=\mathe^{spatial} \cup \mathe^{temporal},\mathr=\{spatial, tempo\})$
    \end{algorithmic}
\end{algorithm}
\subsection{Model Architecture}\label{section:model_arch}
In order to model both the spatial and temporal relations between superpixels, our model primarily relies on the Neural Relational Model \cite{schlichtkrull2018modeling}, which is an extension of GCNs \cite{kipf2016semi} to large-scale relational data. In a Neural Relational Model, the propagation model for calculating the forward-pass update of a node, denoted by $v_i$, is defined as
\begin{equation}\small
    h_{i}^{(l+1)}=\sigma \left(\sum_{r\in \mathr}\sum_{j\in \neighborhood_{i}^{r}}{\frac{1}{c_{i,r}} W_{r}^{(l)}h_{j}^{(l)}+W_{0}^{(l)}h_{i}^{(l)}} \right),
\end{equation}
where $\neighborhood^r_i$ denotes the set of neighbor indices of node $i$ under relation \text{$r\in \mathr$} (not to be confused with the notation $\neighborhood_{t,i}$ from Eq. \ref{eq:neighborhood}). $c_{i,r}$ is a problem-specific normalization constant that can either be learned or chosen in advance (such as \text{$c_{i,r}=|\neighborhood^r_i|)$}. To incorporate edge features, we adapt the approach proposed in \cite{corso2020principal}, that concatenates node and edge attributes as a layer's input, yielding the following:
\begin{equation}\label{eq:concat_edges}\small
    h_{i}^{(l+1)}=\sigma \left(\sum_{r\in \mathr}\sum_{j\in \neighborhood_{i}^{r}}{\frac{1}{c_{i,r}} W_{r}^{(l)}[h_{j}^{(l)},e_{i,j}]+W_{0}^{(l)}h_{i}^{(l)}} \right),
\end{equation}
where $e_{i,j}$ is the feature of the edge connecting nodes \text{$v_i,v_j$}.

\subsection{Augmentations}\label{section:augmentations}
We introduce a few possible augmentations that we found useful for training our model as they improved the generalization.
\paragraph{Additive Gaussian Edge Noise (AGEN) -}
Edge attributes represent distances between superpixel centroids. The coordinates of those centroids may vary due to different superpixel shapes with different centers of mass. To compensate for this, we add a certain amount of noise to each edge attribute. Given a hyper-parameter $\sigma_{edge}$, for each edge attribute $e_{u,v}$ and for each training iteration, we sample a normally distributed variable $z_{u,v}\sim N(0,\sigma_{edge})$ that is added to the edge attribute.

\paragraph{Additive Gaussian Node Noise (AGNN) -}
Node attributes represent the colors of regions in each frame. Similar to edge attributes, the mean color of each segmented region may vary due to different superpixel shapes. To compensate for this, we add a certain amount of noise to each node attribute. Given a hyper-parameter $\sigma_{node}$, for each node attribute $v.c$ of dimension $d_c$ and for each training iteration, we sample a normally distributed variable $z_{v}\sim N_{d_c}(0,\sigma_{node}\cdot I_{d_c})$ that is added to the node attribute.

\paragraph{Random Removal of Spatial Edges (RRSE) -}
This augmentation tends to mimic the regularization effect introduced in DropEdge \cite{rong2019dropedge}. Moreover, since the removal of edges leads to fewer message-passings in a GCN, this also accelerates the training and inference. To perform this, we choose a probability \text{$p_{edge}\in[0,1]$}. Then, each edge $e$ is preserved with a probability of $p_{edge}$.

\paragraph{Random Removal of Superpixels (RRS) -}
SLIC \cite{achanta2010slic} is sensitive to its initialization. Consequently, each video clip may have several graph representations during different training iterations and inference. This can be mitigated by removing a certain amount of superpixels. The outcome is fewer nodes in the corresponding representative graph, as well as fewer edges. Similar to RRSE, we choose a probability \text{$p_{node}\in[0,1]$} so that each superpixel is preserved with a probability of $p_{node}$.

\subsection{Benefits of \textbf{\methodname}}
\paragraph{Invariance -}The absence of coordinates leads to invariance in the spatial dimension. It is evident that such a representation is invariant to rotations and flips since the relations between different parts of the image are solely characterized by distances. This, in turn, obviates the need to perform such augmentations during training.
\paragraph{Efficiency -}We argue that our graph-based representation is more efficient than raw frames. To illustrate this, let $T, C, H$ and $W$ be the dimensions of a clip; that is, the number of frames, number of channels, height and width of a frame, respectively. Correspondingly, the raw representation requires \text{$T\cdot C\cdot H\cdot W$}. To calculate the size of the graph-video, let $S$ be the number of superpixels in a frame. By construction, there are at most \text{$4\cdot S$} edges in each frame because SLIC constraints each to have 4 neighbors. Each edge contains $3$ values, corresponding to the distance on the grid, source and target nodes. Additionally, there are, at most, $S$ edges between every temporal step. This results in \text{$3\cdot (4\cdot S + (T - 1) \cdot S) + C\cdot T\cdot S$} parameters in total. Typically, the second requires much fewer parameters because we choose $S$ so that \text{$S \ll H\cdot W$}.
\paragraph{Prior Knowledge Incorporation -}
Optical-flow and over-segmentation are encoded within the graph-video representation using the inter-frame and intra-frame edges. This incorporates strong prior knowledge within the resultant representation. For example, optical-flow dramatically improved the accuracy in the two-stream methodology that was proposed in \cite{simonyan2014two}. Additionally, over-segmentation using superpixels has been found useful as input features for machine learning models due to the limited loss of important details, accompanied by a dramatic reduction in the expended time by means of reducing the number of elements of the input \cite{proceedings401,dadsetan2021superpixels,avelar2020superpixel}.
\section{Experiments}\label{section:experiments}
We validated \methodname\ on 2 human-action-classification benchmarks. The goal of human action classification is to determine the human-involved action that occurs within a video.
The objectives of this empirical study were twofold:
\begin{itemize}
    \item Analyze the impact of the various parameters on the accuracy of the model.
    \item As we first and foremost target efficiency, we sought to examine the resources' consumption of \methodname\ in terms of Floating Point Operations
    (FLOPs). We followed the conventional protocol \cite{feichtenhofer2020x3d}, which uses single-clip FLOPs as a basic unit of computational cost. We show that we are able to achieve a significant improvement in efficiency over previous methods while preserving state-of-the-art performance.
\end{itemize}
\subsection{Setup}

\paragraph{Datasets -}
We use two common datasets for action classification: \textit{Kinetics-400 (K400)} \cite{kay2017kinetics} and \textit{Charades} \cite{sigurdsson2016hollywood}. Kinetics-400 \cite{kay2017kinetics} is a large-scale video dataset released in 2017 that contains 400 classes, with each category consisting of more than 400 videos. It originally had, in total, around 240K, 19K, and 38K videos for training, validation and testing subsets, respectively. Kinetics is gradually shrinking over time due to videos being taken offline, making it difficult to compare against less recent works. We used a dataset containing 208K, 17K and 33K videos for training, validation and test respectively. We report on the most recently available videos. Each video lasts approximately 10 seconds. The Charades dataset \cite{sigurdsson2016hollywood} is composed of 9,848 videos of daily indoor activities, each of an average length of 30 seconds. In total, the dataset contains 66,500 temporal annotations for 157 action classes. In the standard split, there are 7,986 training videos and 1,863 validation videos, sampled at 12 frames per second. We follow prior arts by reporting the Top-1 and Top-5 recognition accuracy for Kinetics-400 and mean average precision (mAP) for Charades.

\begin{figure}[t]
    \centering
    \includegraphics[width=0.65\linewidth]{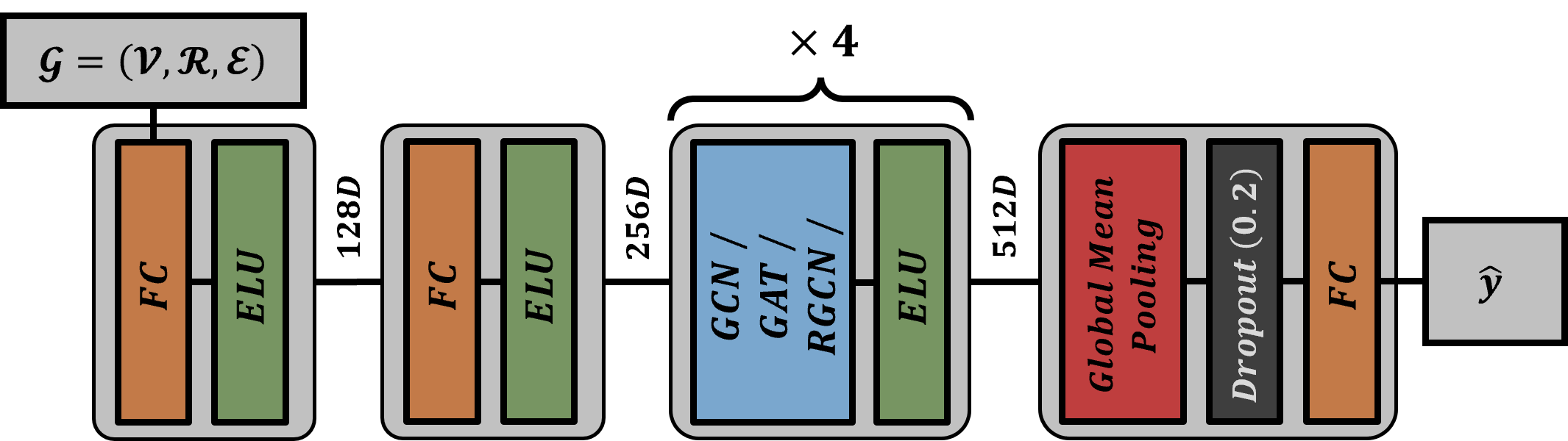}
    \caption{The general graph neural network architecture we use in our experiments.}
    \label{fig:general_arch}
\end{figure}

\paragraph{Network Architecture and Training -}
We use GNN variants and feed each of them with our video-graphs. Specifically, we consider Graph Convolutional Networks \cite{kipf2016semi} (GCNs), Graph Attention Networks \cite{velivckovic2017graph} (GATs) and Relational Graph Convolutional Networks \cite{schlichtkrull2018modeling} (RGCNs). The general architecture of our backbones is depicted in Fig. \ref{fig:general_arch}. It consists of $2$ fully-connected (FC) layers with exponential linear unit (ELU) activations that project the node features into a $256D$ feature space. Then come $4$ layers of the corresponding GNN layer (either GCN, GAT or RGCN along with an edge feature concatenation from Eq. \ref{eq:concat_edges}) with a hidden size of 512 with ELU activations, followed by global mean pooling, dropout with a probability of $0.2$ and a linear layer whose output is the predicted logits. For the GAT layers, we use 4 attention heads in each layer, and average the attention heads' results to obtain the desired hidden layer size. For the RGCN layers, we specify 2 relations, which correspond to the spatial and temporal relations, as described in Section \ref{section:methodology}. We use the Adam \cite{kingma2014adam} with a constant learning rate of \text{$1e-3$} for optimization. While choosing this architecture, the core idea was to keep the architecture simple and shallow, while changing the interaction module to better model the relations between parts of the clip.
We divide the videos into clips using a sliding window of 20 frames, using a stride of 2 between consecutive frames and a stride of 10 between clips. In all the experiments, we used a fixed batch size of 200.

\paragraph{Inference -} 
At the test phase, we use the same sliding window methodology as in the training. We follow the common practice of processing multiple views of a long video and average per-view logits to obtain the final results. The views are drawn uniformly across the temporal dimension of the video, without spatial cropping. The number of views is determined by the validation dataset.

\paragraph{Implementation Details -} All experiments were run on a Ubuntu 18.04 machine with Intel i9-10920X, 93GB RAM and 2 GeForce RTX 3090 GPUs. Our implementation of \methodname\ is in Python3. To generate superpixels, we use \textit{fast-slic} \cite{fastslic} with the AVX2 instruction set. To train the graph neural models, we use Pytorch-Geometric \cite{fey2019fast}. 
We use a fixed seed for SLIC and cache the generated graphs during the first training epochs in order to further reduce the computations. We also store the edge indexes as int16 instead of int64 in order to reduce the memory footprint. Eventually, the memory footprints of the cached datasets is comparable to those of the original ones.

\subsection{Ablation Study}\label{section:ablation}
We conduct an in-depth study on Kinetics-400 to analyze the performance gain contributed by incorporating the different components of \methodname.
\paragraph{Graph Neural Network Variants and Number of Superpixels per Frame -}
We assess the performance of different GNN variants: GCN \cite{kipf2016semi} is trained without edge relations (\ie\, temporal and spatial edges are treated via the same weights). GAT \cite{velivckovic2017graph} is trained by employing the attention mechanism for neighborhood aggregation without edge relations. RGCN \cite{schlichtkrull2018modeling} is trained with edge relations, as described in Section \ref{section:model_arch}.

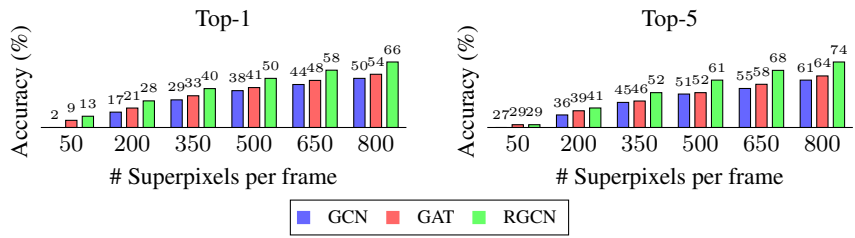
\begin{figure}[!ht]
    \centering
    \begin{subfigure}[b]{0.48\textwidth}
         \centering
         \begin{tikzpicture}
      \begin{axis}[
        height=1.in,
        width=1.1\linewidth,
        axis on top,
        title={Top-1},
        bar width=0.15cm,
        enlarge y limits={value=.1,upper},
        axis x line*=bottom,
        axis y line*=left,
        y axis line style={opacity=0},
        tickwidth=0pt,
        enlarge x limits=true,
        label style={font=\footnotesize},
        ticklabel style = {font=\footnotesize},
        ytick=\empty,
        legend style={
            font=\scriptsize,
            nodes={scale=1, transform shape},
            at={(0.5,-0.8)},
            anchor=south,
            legend columns=-1,
            /tikz/every even column/.append style={column sep=0.5cm}
        },
        xlabel={\# Superpixels per frame},
        ylabel={Accuracy (\%)},
        xtick={50,200,350,500,650,800},
        nodes near coords,
       nodes near coords style={font=\tiny},
       ybar = 2.0pt
    ]
    \addplot [fill=blue!60] coordinates {
(50, 2)
(200, 17) 
(350, 29)
(500, 38)
(650, 44)
(800, 50)
      }; 
   \addplot [fill=red!60] coordinates {
(50, 9)
(200, 21)
(350, 33)
(500, 41)
(650, 48)
(800, 54)
      }; 
   \addplot [fill=green!60] coordinates {
(50, 13)
(200, 28)
(350, 40)
(500, 50)
(650, 58)
(800, 66)
      }; 

  \end{axis}
\end{tikzpicture}
     \end{subfigure}
    \begin{subfigure}[b]{0.48\textwidth}
         \centering
         \begin{tikzpicture}
      \begin{axis}[
        height=1.in,
        width=1.1\linewidth,
        axis on top,
        title={Top-5},
        bar width=0.15cm,
        enlarge y limits={value=.1,upper},
        axis x line*=bottom,
        axis y line*=left,
        y axis line style={opacity=0},
        tickwidth=0pt,
        enlarge x limits=true,
        label style={font=\footnotesize},
        ticklabel style = {font=\footnotesize},
        legend style={
            font=\scriptsize,
            nodes={scale=1, transform shape},
            at={(0.5,-0.8)},
            anchor=south,
            legend columns=-1,
            /tikz/every even column/.append style={column sep=0.5cm}
        },
        ytick=\empty,
        xlabel={\# Superpixels per frame},
        ylabel={Accuracy (\%)},
        xtick={50,200,350,500,650,800},
        nodes near coords,
       nodes near coords style={font=\tiny},
       ybar = 2.0pt
    ]
    \addplot [fill=blue!60] coordinates {
(50, 27)
(200, 36) 
(350, 45)
(500, 51)
(650, 55)
(800, 61)
      };
   \addplot [fill=red!60] coordinates {
(50, 29)
(200, 39)
(350, 46)
(500, 52)
(650, 58)
(800, 64)
      };
   \addplot [fill=green!60] coordinates {
(50, 29)
(200, 41)
(350, 52)
(500, 61)
(650, 68)
(800, 74)
      };
  \end{axis}
\end{tikzpicture}
     \end{subfigure}

    \begin{tikzpicture}
    \begin{axis}[
        axis lines=none,
        height=0.7in,
        width=0.8in,
        title={},
        xmin=0, xmax=1,
        ymin=0, ymax=1,
        ytick={},
        xtick={},
        legend columns=3,
        legend style={anchor=south east,at={(1,0)},nodes={scale=0.8, transform shape}, column sep=0.2cm},
    ]
     \addplot+[forget plot]
       coordinates{
   (0, 2.388854806)
       };
    \addlegendimage{color=blue!60,mark=square*,style=only marks,line width=1pt}
    \addlegendentry{GCN}
    \addlegendimage{color=red!60,mark=square*,style=only marks,line width=1pt}
    \addlegendentry{GAT}
    \addlegendimage{color=green!60,mark=square*,style=only marks,line width=1pt}
    \addlegendentry{RGCN}
    \end{axis}
    \end{tikzpicture}
    \caption{The effect of varying the number of superpixels on test accuracy on K-400.}
    \label{fig:n_sp_and_model_variants_ablation}
\end{figure}

The results of the action classification on K-400 are shown in Figure \ref{fig:n_sp_and_model_variants_ablation}. In this series, the number of views is fixed at $8$, which is the number of views that was found to be most effective for the validation set. For all variants, increasing the number of superpixels per frame ($S$) contributes to the accuracy. We notice a significant improvement in accuracy for the lower range of the number of superpixels, while the accuracy begins to saturate for \text{$S\geq 650$}. Increasing further the number of superpixels leads to bigger inputs, which require more computations. As our goal is to maximize the efficiency, we do not experiment with larger inputs in this section. 

We further present in Table \ref{table:models_ablation} the models' specifications for $800$ superpixels, which is the best-performing configuration in this series of experiments. Unsurprisingly, the GCN variant requires the least amount of computations. Meanwhile, the RGCN variant requires fewer computations than GAT and achieves a higher level of accuracy. We conclude that it is beneficial to incorporate edge relations when wishing to encode temporal and spatial relations in videos, and that those features are not easily learned by heavy computational models, such as GAT.

\begin{table}[!ht]
\caption{Comparison of model specifications for various architectures. We report the Top-1 and Top-5 accuracy on Kinetics-400.} 
\label{table:models_ablation}
\centering
\begin{tabularx}{\linewidth}{X X X X X X}
\hline
Model & Top-1 & Top-5 & FLOPs ($\cdot10^9$) & Params ($\cdot10^6$)\\
\hline
$GCN$ & 50.1 & 61.6 & 28 & 2.08\\
$GAT$ & 54.7 & 64.5 & 56 & 3.93\\
$RGCN$ & 66.2 & 74.1 & 42 & 2.99\\
\hline
\end{tabularx}
\end{table}

\paragraph{Augmentations -}
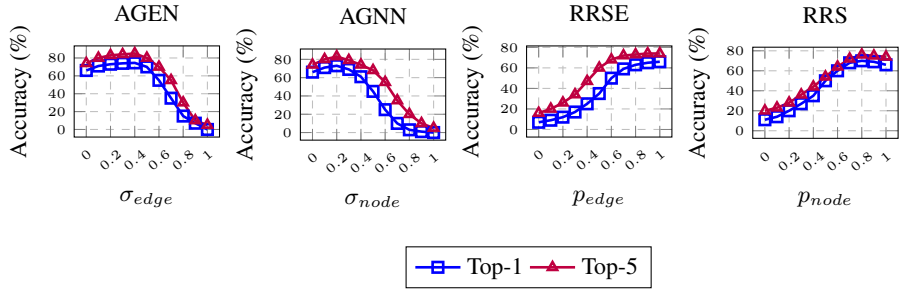
\begin{figure*}[t]
    \centering
    \begin{subfigure}[b]{0.24\textwidth}
         \centering
         \begin{tikzpicture}
    \begin{axis}[
        height=1.1in,
        width=1.2\linewidth,
        title={AGEN},
        xlabel={$\sigma_{edge}$},
        ylabel={Accuracy (\%)},
        xmin=0, xmax=1,
        xtick={0,0.2, 0.4, 0.6, 0.8, 1},
        xticklabel style={font=\tiny, rotate=40},
        yticklabel style={font=\tiny},
        ymajorgrids=true,
        xmajorgrids=true,
        grid style=dashed,
        enlargelimits=0.1,
        legend pos=south west,
        label style={font=\footnotesize},
    ]
     
    \addplot[
        color=blue,mark=square,style=solid,line width=1pt
        ]
        coordinates {
(0, 66)
(0.1, 71)
(0.2, 73)
(0.3, 74)
(0.4, 74.5)
(0.5, 70)
(0.6, 55)
(0.7, 35)
(0.8, 15)
(0.9, 7)
(1.0, 0.0025)
        }; 
        
        \addplot[
        color=purple,mark=triangle,style=solid,line width=1pt
        ]
        coordinates {
(0, 74)
(0.1, 80)
(0.2, 83)
(0.3, 84)
(0.4, 85)
(0.5, 80)
(0.6, 70)
(0.7, 55)
(0.8, 30)
(0.9, 10)
(1.0, 5)
        };
 
    \end{axis}
    \end{tikzpicture}
         \label{fig:ablation_agen}
     \end{subfigure}
    \begin{subfigure}[b]{0.24\textwidth}
         \centering
         \begin{tikzpicture}
    \begin{axis}[
        height=1.1in,
        width=1.2\linewidth,
        title={AGNN},
        xlabel={$\sigma_{node}$},
        ylabel={Accuracy (\%)},
        xmin=0, xmax=1,
        xtick={0,0.2, 0.4, 0.6, 0.8, 1},
        xticklabel style={font=\tiny, rotate=40},
        yticklabel style={font=\tiny},
        ymajorgrids=true,
        xmajorgrids=true,
        grid style=dashed,
        enlargelimits=0.1,
        legend pos=south west,
        label style={font=\footnotesize},
    ]
     
    \addplot[
        color=blue,mark=square,style=solid,line width=1pt
        ]
        coordinates {
(0, 66)
(0.1, 71)
(0.2, 73)
(0.3, 69)
(0.4, 61)
(0.5, 45)
(0.6, 25)
(0.7, 10)
(0.8, 3)
(0.9, 1)
(1.0, 0.0025)
        }; 
        
        \addplot[
        color=purple,mark=triangle,style=solid,line width=1pt
        ]
        coordinates {
(0, 74)
(0.1, 80)
(0.2, 83)
(0.3, 79)
(0.4, 74)
(0.5, 68)
(0.6, 55)
(0.7, 35)
(0.8, 20)
(0.9, 10)
(1.0, 5)
        };
 
    \end{axis}
    \end{tikzpicture}
         \label{fig:ablation_agnn}
     \end{subfigure}
    \begin{subfigure}[b]{0.24\textwidth}
         \centering
         \begin{tikzpicture}
    \begin{axis}[
        height=1.1in,
        width=1.2\linewidth,
        title={RRSE},
        xlabel={$p_{edge}$},
        ylabel={Accuracy (\%)},
        xmin=0, xmax=1,
        ymin=0, 
        xtick={0,0.2, 0.4, 0.6, 0.8, 1},
        xticklabel style={font=\tiny, rotate=40},
        yticklabel style={font=\tiny},
        ymajorgrids=true,
        xmajorgrids=true,
        grid style=dashed,
        enlargelimits=0.1,
        legend pos=south west,
        label style={font=\footnotesize},
    ]
     
    \addplot[
        color=blue,mark=square,style=solid,line width=1pt
        ]
        coordinates {
(0, 7)
(0.1, 9)
(0.2, 12)
(0.3, 17)
(0.4, 25)
(0.5, 35)
(0.6, 50)
(0.7, 59)
(0.8, 63)
(0.9, 65)
(1.0, 66)
        }; 
        
        \addplot[
        color=purple,mark=triangle,style=solid,line width=1pt
        ]
        coordinates {
(0, 16)
(0.1, 20)
(0.2, 26)
(0.3, 34)
(0.4, 47)
(0.5, 60)
(0.6, 68)
(0.7, 72)
(0.8, 73)
(0.9, 74)
(1.0, 74)
        };
 
    \end{axis}
    \end{tikzpicture}
         \label{fig:ablation_rrpe}
     \end{subfigure}
    \begin{subfigure}[b]{0.24\textwidth}
         \centering
         \begin{tikzpicture}
    \begin{axis}[
        height=1.1in,
        width=1.2\linewidth,
        title={RRS},
        xlabel={$p_{node}$},
        ylabel={Accuracy (\%)},
        xmin=0, xmax=1,
        ymin=0,
        xtick={0,0.2, 0.4, 0.6, 0.8, 1},
        xticklabel style={font=\tiny, rotate=40},
        yticklabel style={font=\tiny},
        ymajorgrids=true,
        xmajorgrids=true,
        grid style=dashed,
        enlargelimits=0.1,
        legend pos=south west,
        label style={font=\footnotesize},
    ]
     
    \addplot[
        color=blue,mark=square,style=solid,line width=1pt
        ]
        coordinates {
(0, 11)
(0.1, 14)
(0.2, 20)
(0.3, 27)
(0.4, 35)
(0.5, 50)
(0.6, 60)
(0.7, 68)
(0.8, 70)
(0.9, 69)
(1.0, 66)
        }; 
        
        \addplot[
        color=purple,mark=triangle,style=solid,line width=1pt
        ]
        coordinates {
(0, 20)
(0.1, 23)
(0.2, 28)
(0.3, 36)
(0.4, 44)
(0.5, 54)
(0.6, 64)
(0.7, 72)
(0.8, 76)
(0.9, 75)
(1.0, 74)
        };
 
    \end{axis}
    \end{tikzpicture}
         \label{fig:ablation_rrp}
     \end{subfigure}
     
     \begin{subfigure}[b]{0.1\textwidth}
         \centering
         \begin{tikzpicture}
    \begin{axis}[
        axis lines=none,
        height=0.8in,
        width=0.8in,
        title={},
        xmin=0, xmax=1,
        ymin=0, ymax=1,
        ytick={},
        xtick={},
        legend columns=2,
        legend style={anchor=south, at={(0,0),nodes={scale=0.8, transform shape}, column sep=0.2cm}}
    ]
     \addplot+[forget plot]
       coordinates{
   (0, 2.388854806)
       };
    \addlegendimage{color=blue,mark=square,style=solid,line width=1pt}
    \addlegendentry{Top-1}
    \addlegendimage{color=purple,mark=triangle,style=solid,line width=1pt}
    \addlegendentry{Top-5}
    \end{axis}
    \end{tikzpicture}
     \end{subfigure}

    \caption{The impact of the proposed augmentations on test accuracy of Kinetics-400: Additive Gaussian edge noise (AGEN). Additive Gaussian node noise (AGNN). Random removal of spatial edges (RRSE). Random removal of superpixels (RRS).}
    \label{fig:augmentations_grid}
\end{figure*}

We assessed the impact of augmentations on the performance and their ability to alleviate over-fitting. For this purpose, we chose the best configuration obtained from the previous experiments, that is, RGCN with 800 superpixels per frame, and trained it while adding one augmentation at a time. The results of this series are depicted in Figure \ref{fig:augmentations_grid}. Each graph shows the level of accuracy reached by training the model with one of the parameters that control the augmentation.

We begin with the analysis of the AGEN and AGNN, both relate to the addition of Gaussian noise to the graph components, with the corresponding parameters controlling the standard deviations. Their impact is unnoticeable as these parameters head towards $0$, since lower values reflect the scenarios in which little or no augmentations are applied. Slightly increasing the parameter brings about a gradual improvement in the accuracy, until a turning point is reached, after which the level of accuracy declines until it reaches \text{$\sim \frac{1}{400}$}, which resembles a random classifier. The decrease in accuracy stems from the noise obscuring the original signal, allegedly forcing the classifier to classify ungeneralizable noise.
For RRSE and RRS, the random removal of spatial edges harms the accuracy of the model. This finding leads us to conclude that spatial edges encode meaningful information about relations between the entities. Moreover, slightly removing the nodes contributes to the level of accuracy, reaching a peak at \text{$p_{node}\approx 0.8$}. To conclude, we present the values that lead to the best Top-1 accuracy score in Table \ref{table:augmentations_params}.
\begin{table}[!h]
\caption{Augmentation parameters and their optimized values.} 
\label{table:augmentations_params}
\centering
\begin{tabularx}{\linewidth}{X X X X X}
\hline
Param & $\sigma_{edge}$ & $\sigma_{node}$ & $p_{edge}$ & $p_{node}$ \\
\hline
Value & 0.4 & 0.2 & 1 & 0.8\\
Top-1 & 74.5 & 73 & 66 & 70\\
Top-5 & 85 & 83 & 74 & 76\\
\hline
\end{tabularx}
\end{table}
\subsection{Comparison to the State-of-the-Art}
\begin{figure*}[t]
    \centering
    \begin{subfigure}[b]{0.49\textwidth}
         \centering
         \includegraphics[width=\linewidth]{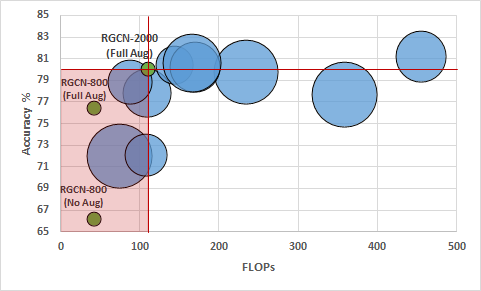}
         \caption{FLOPS vs Kinetics-400 Accuracy}
         \label{fig:k400_relative_sota}
     \end{subfigure}
    \begin{subfigure}[b]{0.49\textwidth}
         \centering
         \includegraphics[width=\linewidth]{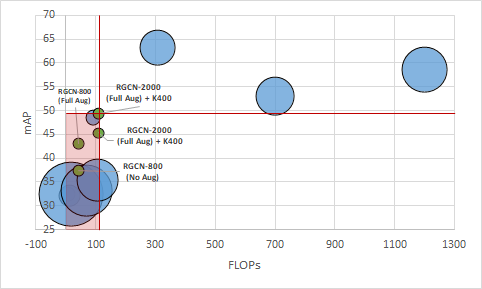}
         \caption{FLOPS vs Charades mAP}
         \label{fig:charades_relative_sota}
     \end{subfigure}
    \caption{\textbf{Model FLOPs vs. performance - } Green bubbles indicates \methodname\ variants, radius indicates the number of parameters. To avoid overloading, identities of the other models are omitted. For both datasets, RGCN-2000 with the full set of augmentations is on par with the state-of-the-art, while greatly reducing model size and FLOPs.}
    \label{fig:augmentations_grid}
\end{figure*}

\paragraph{Kinetics-400 -}
We present the K-400 results for our RGCN variant in Table \ref{table:k400_sota} and Figure \ref{fig:k400_relative_sota}, along with comparisons to previous arts, including convolutional-based and transformer-based methods. Our results are denoted RGCN-$d$, where $d$ represents the number of superpixels. Additionally, we use the set of augmentations with the parameters from Table \ref{table:augmentations_params}. First, when the RGCN-800 model is trained with the full set of augmentations (denoted Full-Aug), it achieves a significantly higher Top-1 accuracy than when it is trained without any augmentation (denoted No-Aug) or when each augmentation is applied individually. These results demonstrate the effectiveness of our model and that our augmentations can alleviate overfitting and improve the generalization over the test set. Second, all our RGCNs require orders-of-magnitude fewer computations than the previous arts, as well as more than \text{$\times 10$} fewer parameters.
\begin{table}[h]
\caption{Comparisons to state-of-the-art on the K-400 dataset. We report the Top-1 and Top-5 accuracies. The top section of the table depicts convolution-based models, The middle section depicts transformer-based models, and the bottom section depicts ours.} 
\label{table:k400_sota}
\centering
\begin{tabularx}{\linewidth}{p{3.5cm} X X X p{2cm} p{2cm}}
\hline
Method & Top-1 & Top-5 & Views & FLOPs ($\cdot10^9$) & Param ($\cdot10^6$)\\
\hline 
SlowFast R101+N \cite{feichtenhofer2019slowfast} & 79.8 & 93.9 & 30 & 234 & 59.9\\
X3D-XXL R101+N \cite{feichtenhofer2020x3d} & 80.4 & 94.6 & 30 & 144 & 20.3\\

\hline
MViT-B, 32×3 \cite{fan2021multiscale}& 80.2 & 94.4 & 5 & 170 & 36.6\\
TimeSformer-L \cite{bertasius2021space}& 80.7 & 94.7 & 3 & 2380 & 121.4\\
ViT-B-VTN \cite{neimark2021video}& 78.6 & 93.7 & 1 & 4218 & 11.04\\
ViViT-L/16x2 \cite{arnab2021vivit}& 80.6 & 94.7 & 12 & 1446 & 310.8\\
Swin-S \cite{liu2021video}& 80.6 & 94.5 & 12 & 166 & 49.8\\
\hline
RGCN-800 (No / Full Aug) & 66.2 / 76.4 & 74.1 / 91.1 & 8 & \textbf{42} & \textbf{2.57}\\
RGCN-2000 (Full Aug) & 80.0 & 94.3 & 8 & \textbf{110} & \textbf{2.57}\\
\hline
\end{tabularx}
\end{table}
\paragraph{Charades -}
We train RGCN variants with $800$ and $2000$ superpixels with the set of augmentations found in Table \ref{table:augmentations_params}. We also follow prior arts \cite{feichtenhofer2019slowfast,fan2021multiscale} by pre-training on K-400 followed by replacing the last FC layer and fine-tuning on Charades. Table \ref{table:charades_sota} and Figure \ref{fig:charades_relative_sota} show that when our RGCN model is trained with 2000 superpixels, its mAP score is comparable to the current state-of-the-art, but this score is reached with orders-of-magnitude fewer computations and using considerably fewer parameters. 
\begin{table}[h]
\caption{Comparisons to state-of-the-art on the Charades multi-label dataset. We report the mAP scores as more than one ground truth action is possible.} 
\label{table:charades_sota}
\centering
\begin{tabularx}{\linewidth}{p{5cm} X X X}
\hline
Method & mAP & FLOPs ($\cdot10^9$) & Params ($\cdot10^6$)\\
\hline
MoVieNet-A2 \cite{kondratyuk2021movinets} & 32.5 & 6.59 & 4.8\\
MoVieNet-A4 \cite{kondratyuk2021movinets} & 48.5 & 90.4 & 4.9\\
\hline
TVN-1 \cite{piergiovanni2019tiny} & 32.2 & 13 & 11.1\\
TVN-4 \cite{piergiovanni2019tiny} & 35.4 & 106 & 44.2\\
\hline
AssembleNet-50 \cite{ryoo2019assemblenet} & 53.0 & 700 & 37.3\\
AssembleNet-101 \cite{ryoo2019assemblenet} & 58.6 & 1200 & 53.3\\
\hline
SlowFast \text{$16\times 8$} R101 \cite{feichtenhofer2019slowfast} & 45.2 & 7020 & 59.9\\
\hline
RGCN-800 (No Aug/Full Aug) & 37.4 / 43.1 & \textbf{42} & \textbf{2.57}\\
RGCN-2000 (Full Aug/+K400) & 45.3 / 49.4 & \textbf{110} & \textbf{2.57}\\
\hline
\end{tabularx}
\end{table}

\subsection{Video-Graph Generation Run-Time}
\begin{wrapfigure}[15]{r}{0.5\linewidth}
\begin{center}
\begin{tikzpicture}
    \begin{axis}[
        height=1.6in,
        width=\linewidth,
        xlabel={Number of superpixels},
        ylabel={Time (seconds)},
        xtick={400,800,1200,1600,2000},
        xticklabel style={rotate=40},
        ymajorgrids=true,
        xmajorgrids=true,
        grid style=dashed,
        enlargelimits=0.05,
        legend pos=south west,
    ]
     
    \addplot[
    color=black,mark=square,style=solid,line width=1pt
    ]
    coordinates {
    (200, 0.03541163465806415)
    (400, 0.03740545642984156)
    (600, 0.04594675779342651)
    (800, 0.05070319216875802)
    (1000, 0.0571724717957633)
    (1200, 0.0609033953824213)
    (1400, 0.0712082170695066)
    (1600, 0.0752192445859617)
    (1800, 0.0818014575434856)
    (2000, 0.0934623439846005)
    };
    \end{axis}
    \end{tikzpicture}
\end{center}
\caption{Time of generation depending on the number of superpixels.}
\label{fig:graph_run_time}
\end{wrapfigure}
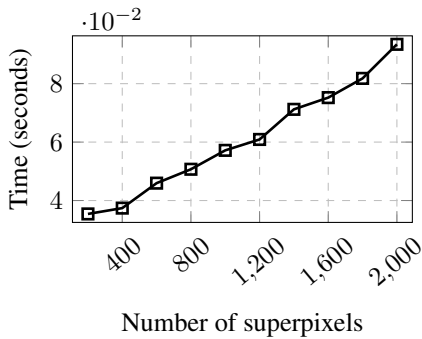
The transition into a video-graph representation requires the consideration of the time needed for generating it. In Figure \ref{fig:graph_run_time}, we measured the average time needed using our setup, which include the whole pipeline: \textbf{1.} Superpixels calculation, and \textbf{2.} Graph structure generation, that is, creating edges between adjacent super-pixels and features calculation as described in Section \ref{section:methodology}. Interestingly, the first step is relatively short compared to the second. Apparently, the optimized \textit{fast-slic} \cite{fastslic} performs well, while the search for adjacent superpixels is time consuming. This opens the possibilities of further optimization.

\section{Conclusions and Future Work}\label{section:conclusions}
In this paper, we present \methodname, a graph video representations that enable video-processing via graph neural networks. Furthermore, we propose a relational graph convolutional model that suits this representation. Our experimental study demonstrates this model's efficiency in performing video-related tasks while achieving comparable performance to the current state-of-the-art. An interesting avenue for future work is to explore new graph representations of videos, including learnable methods. Additionally, we consider the development of new dedicated graph neural models for processing the unique and dynamic structure of the video-graph as an interesting research direction. Finally, unified models for image and video understanding that disregard temporal edges could be explored in order to take advantage of the amount of data in both worlds.

\clearpage
%
%
\bibliographystyle{splncs04}
\bibliography{egbib}
\end{document}